# Adaptive Generation Model: A New Ensemble Method

Jiacheng Ruan[1], Jiahao Li[2]

[1]SAILAB, Xidian University, Xifeng Road, Xi'an, China
[2]Department of Merchant Shipping, Shanghai Maritime University,Shanghaii, China
{ jcruan, 201930110137}@stu.xidian.edu.cn, @stu.shmtu.edu.cn

Abstract: As a common method in Machine Learning, Ensemble Method is used to train multiple models from a data set and obtain better results through certain combination strategies. Stacking method, as representatives of Ensemble Learning methods, is often used in Machine Learning Competitions such as Kaggle. This paper proposes a variant of Stacking Model based on the idea of gcForest, namely Adaptive Generation Model (AGM). It means that the adaptive generation is performed not only in the horizontal direction to expand the width of each layer model, but also in the vertical direction to expand the depth of the model. For base models of AGM, they all come from preset basic Machine Learning Models. In addition, a feature augmentation method is added between layers to further improve the overall accuracy of the model. Finally, through comparative experiments on 7 data sets, the results show that the accuracy of AGM are better than its previous models.

## 1 INTRODUCTION

Ensemble learning [Zhou et al., 2012] is a powerful combined learning method. It trains multiple weak learners and then combines them through different strategies to obtain a strong learner, improving the overall accuracy of the model. Generally speaking, Ensemble Learning methods can be divided into three main types: Bagging [Breiman et al., 1996], Boosting [Schapire et al., 1990] and Stacking [Wolpert et al., 1992]. Among them, there are many epoch-making classic algorithms, such as Bagging method represented by Random Forest [Breiman et al., 2001], and Boosting method represented by XGBoost [Chen et al., 2016].

Stacking method was first proposed by Wolpert, which refers to training a model to combine other models. It means to train multiple different models first, and then use the output of these models as input to train a model to get the final result. However, there are generally only two layers in traditional stacking method. As shown in Figure 1, XGBoost is used in the first layer and the new input features are obtained through 5-fold cross-validation, and then they will be input to the next layer. In the second layer, Logistic Regression will be trained by these new features. Finally, the result was predicted.

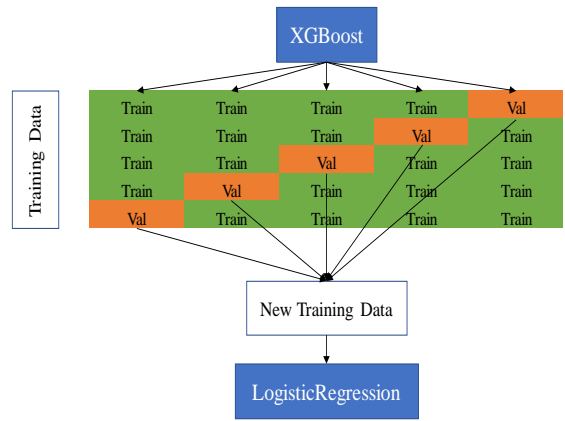

Figure 1: Traditional Stacking Method

However, with the development of DNNs, it is well known that the performance of DNNs is closely related to its depth. With the relative deepening of the depth, it is wonderful that the more features will be extracted and the effect of the model will be greater. Therefore, the idea of multi-layer could be applied to Stacking to increase the depth and enhance the effect of the model. In 2017, Professor Zhou. proposed gcForest [Zhou, 2017], which strengthened the depth of the model and improved the overall model effect through multi-layer cascade. After the experiment, it also proved that the effect of multi-layer forest was much higher than Two-tier stacking.

In this paper, we implement adaptive generation not only on its depth, but also on the width of each layer. Whereby we propose the Adaptive Generation Model (AGM),which can adaptively generate using different input data.

Through cascading in each layer, the width of the model is adaptively increased to improve the performance of the model in the current layer; By using cascading method in the vertical direction, the depth of the model is increased to improve the overall effect of the model. In addition, when delivering feature between layers, we introduce a feature enhancement method--PCA to rotate the feature space [Zhang and Zhang et al., 2008].

In order to verify the effect of the model, we report experiments in Chapter 4. We conduct ablation experiments about all three versions of AGM on 7 classification datasets. In addition, we also conduct comparative experiments with gcForest, RandomForest and XGBoost algorithms to further prove a result that AGM algorithm can improve the overall accuracy of the model.

## 2 RELATED WORK

### 2.1 The Idea of Adaptive Generation

Professor Zhou. proposed gcForest in 2017 which applied the hierarchical structure to Random Forests. In deep neural networks, representation learning mainly relied on the layer-by-layer processing of original features. Inspired by this idea, he adopted a cascade structure for gcForest in which each cascade receives the feature information processed by its previous layer, and outputs its processing results to the next layer. In addition, gcForest gives the idea of adaptive generation. Namely, after generating the next level, it would estimate the performance of the entire cascade model on the validation set. If there is no significant performance gain, the training process will be terminated to automatically determine the growth of the model.

But in gcForest, the number of models in each layer are pre-set parameters, and the model as a whole is only deepening in its own depth. This gives us inspiration, AGM not only adaptively generates in depth, but also in width ,seen Chapter 3 for specific methods.

### 2.2 Feature Augmentation Method

AugBoost [Tannor et al., 2019] proposed by Philip Tannor et al. in 2019, based on GBDT, added different methods to improve the performance of GBDT. Namely, an extra training Artificial Neural Network was used between iterations of the GBDT to extract the features from last hidden layer, and the extracted features will be concatenated to the original feature for feature augmentation, or used PCA and RP methods to rotate the feature-space for feature augmentation.

On the contrary, in gcForest, there is no similar Feature Augmentation Methods between layers, but just feature concatenating. Therefore, for AGM, we have added the method of feature augmentation between layers to improve the performance of the model better.

## 3 METHODS

### 3.1 Adaptive Generation in Width

In gcForest, the number of models in each layer is determined, while in AGM, the number of models in each layer is adaptively generated. The purpose of adaptive generation in width is to determine the number of models in this layer which means the number of new features is input to the next layer. As shown in the figure 2, $X\_train_{n-1}, Y\_train$ represents the feature matrix and target vector of the training set from the previous layer to this layer, $X\_val_{n-1}, Y\_val$ represents the feature matrix and target vector of validation set from the previous layer to this layer. The training set will be used as input to $model_1$ for training, and then use the trained $model_1$ to predict the training set and the test set respectively to get $Y\_train\_predict1$ and $Y\_val\_predict1$. Concatenate the prediction result with $X\_train_{n-1}$ and $X\_val_{n-1}$ to get $X\_train1$ and $X\_val1$, which will be used as the input to the test model for training and evaluation.

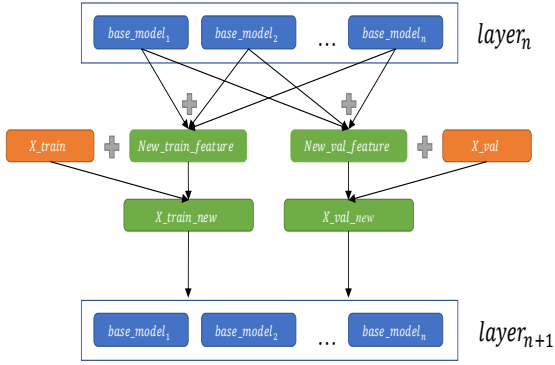

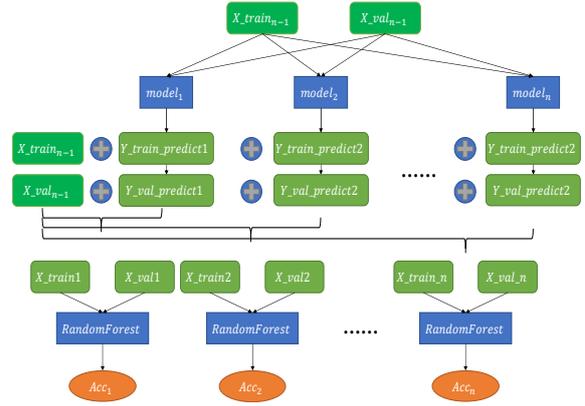

Figure 2: The Method of Performing Adaptive Generation in width

Figure 3: The Method of Performing Adaptive Generation in depth

In the same way, the training set and the validation set will be used as the input to $model_2$ for training and make predictions on the train set and test set to get $Y\_train\_predict2$ and $Y\_val\_predict2$. The prediction results are concatenated with $X\_train1$ and $X\_val1$ to obtain $X\_train2$ and $X\_val2$ for training and evaluation of the test model. If the accuracy of the nth verification is less than the accuracy of the n-1th verification, the growth will be stopped in width. (The test model used in this paper is RandomForest and its purpose is to determine the number of new features through the accuracy of the test)

## 3.2 Adaptive Generation in Depth

For the adaptive generation in the depth direction, we used a cascade method for multi-layer stacking. The model will not stop growing in depth until the new layer could not improve the overall accuracy of the model. As shown in the figure 3, after the training set is used as the input to each base model for training, we used the trained model to predict the training set and the validation set respectively, so we could obtain new feature information by concatenating the original features with the predicted features. Next, the new feature information will be input to the next layer models. In addition to the simple concatenating feature vector operation, we also add a Feature Augmentation Method---PCA.

## 3.3 Feature Augmentation between layers (PCA)

Inspired by the idea of AugBoost, we add a Feature Augmentation Method to the AGM. When the features are spliced between layers, we use an Unsupervised Method—PCA to rotate the feature space. In order to make the input features between layers diversified, we set the value of dimensionality reduction in PCA, which means that the value range of the number of retained features is half of the original feature number to the original feature number.

## 3.4 Pseudo Code

In Algorithm1, we create the pseudo code about how to train a AGM_v3 model. Among the inputs, Model M and val-model H are used to determine the width of each layer. For example, we use Random Forest as Model M and val-model H for AGM_v3 in our experiment. Lines 2-14 shows the function about how to determine the width of each layer. Lines 16-46 shows the process of adaptive generating in depth. In lines 20-22, we concatenate the features processed by its preceding layer and the new features predicted by the models in this layer. Lines 31-44 shows the criterion about how to stop the adaptive generating in depth. After the whole process above, we can obtain AGM_v3 model. Please see appendix for pseudo code

## 4 EXPERIMENT

### 4.1 Experiment process and Results

We use 7 classification data sets [Alcala-Fdez et al., 2011, Dheeru and Karra Taniskidou et al., 2017] for experiments. The data sets are divided into training set and test set according to the ratio of 8 : 2. The detailed information of the data sets is shown in Table 1 below:

Table 1: The details of seven classification data sets

| DataSet | Samples | Features | Classes | Class Proportion |
|---|---|---|---|---|
| Mnist | 1797 | 64 | 10 | 178/182/177/183/181/182/181/179/174/180 |
| Car Evaluation | 1727 | 6 | 4 | 1209/384/69/65 |
| Cardio | 70000 | 11 | 2 | 35021/34979 |
| Cortex | 1080 | 77 | 8 | 150/150/135/135/135/135/135/105 |
| Diabetes | 768 | 8 | 2 | 500/268 |
| Frogs-MFCCs | 7195 | 24 | 10 | 672/542/3478/310/472/1121/270/114/68/148 |
| Gender-by-voice | 3168 | 20 | 2 | 1584/1584 |

1) Experiment 1: Ablation experiment on three versions of AGM

There are three versions of AGM. In the AGM_v1 version, the horizontal width we set is 4, which means there are only 4 models in each layer, so that AGM_v1 does not have the function of adaptive generation in width. Except, between the vertical layers, we do not implement PCA to change the feature space.

In the AGM_v2 version, compared with AGM_v1, PCA is added between layers for feature augmentation. In the AGM_v3 version, we do not set a fixed horizontal width, but add horizontal adaptive generation. The experimental results are shown in Table 2 below:

Table 2: The experimental results among v1, v2 and v3 of AGM

| DataSet | AGM_v1 | AGM_v2 | AGM_v3 |
|---|---|---|---|
| Mnist | 97.22% | 98.05% | **98.33%** |
| Car Evaluation | 97.68% | 97.98% | **98.55%** |
| Cardio | 75.00% | **76.75%** | 75.00% |
| Cortex | 98.14% | 99.07% | **99.54%** |
| Diabetes | 82.46% | 83.76% | **84.42%** |
| Frogs-MFCCs | 99.72% | 99.86% | **99.93%** |
| Gender-by-voice | **98.73%** | 98.42% | 98.58% |

2) Experiment 2: Comparative experiment with gcForest, RandomForest and XGBoost

We use the AGM_v3 model and the three models of gcForest, RandomForest, and XGBoost for comparative experiments. For gcForest, we do not use Multi-Grained Scanning, but use the cascading forest at the back. In addition, the base models used by gcForest is the same as AGM_v3, which are RandomForestClassifier (100Trees), XGBClassifier (100Trees) and ExtraTreesClassifier (100Trees). For RandomForest and XGBoost, we have set 1000Trees. The experimental results are shown in the following Table 3:

Table 3: The experimental results among AGM_v3, RandomForest, XGBoost and gcForest

| DataSet | AGM_v3 | RandomForest(1000Trees) | XGBoost(1000Trees) | gcForest |
|---|---|---|---|---|
| Mnist | **98.33%** | 97.50% | 95.83% | **98.33%** |
| Car Evaluation | **98.55%** | 97.39% | **98.55%** | 98.26% |
| Cardio | 75.00% | 74.00% | 71.25% | **75.00%** |
| Cortex | **99.54%** | **99.54%** | 93.51% | 98.14% |
| Diabetes | **84.42%** | 81.17% | 79.22% | 79.87% |
| Frogs-MFCCs | **99.93%** | 99.65% | 99.65% | 99.65% |
| Gender-by-voice | **98.58%** | 98.26% | 97.63% | **98.58%** |

### 4.2 Discussion and Analysis

In experiment 1, the result of AGM_v3 is better than that of AGM_v2 and the accuracy of v3 on six data sets(except for Cardio) is 0.37% averagely higher than that of v2. The difference between v2 and v3 version is that the latter adds adaptive generation in width, so it can prove that the adaptive generation in width is effective, and make the model better to perform self-adaption on differently distributed data. For AGM_v2 and AGM_v1, the effect of v2 is also better than that of v1 and the accuracy of v2 on 7 data sets is 0.7%,which is averagely higher than that of v1. It can be seen that adding PCA between layers to rotate the feature space for feature augmentation is helpful to improve the overall effect of the model.

In experiment 2, we compared the accuracy of gcForest, RandomForest, XGBoost and AGM_v3 on 7 data sets. From the experimental results, it can be seen that the accuracy of v3 on 7 data sets is 0.98%, 2.67% and 0.93%, which are respectively higher than that of RandomForest, XGBoost and gcForest. Compared with Bagging method of RandomForest and Boosting method of XGBoost, the multi-layer cascade method adopted by AGM_v3 and gcForest is more effective. Compared with gcForest, because not only performs the adaptive generation in depth, but also performs the adaptive generation in width, the effect of AGM_v3 is better than gcForest without multi-grain scanning, which proves performing adaptive generation in width is more helpful for the model to adapt to different distributed data sets.

## 5 CONCLUSIONS AND FUTURE WORK

In this paper, we proposed an adaptive generative model based on the ideas of gcForest and AugBoost. AGM could generate in width and depth, and PCA could be added between layers to rotate the feature space. In the comparison experiments, it can be seen that the effect of adaptive generation both in the horizontal and vertical directions is better than that of only in the vertical direction. So performing adaptive generation in two directions can better adapt to different types of data.

For further in-depth researches, the AGM model may be expanded and improved from three aspects. Before training the model, we can add Multi-Grained Scanning for feature extraction. Between layers, we can not only use PCA methods for feature augmentation, but also use ANN and other methods for feature augmentation. For the combination strategy, AGM in this paper only use the Voting method for classification problems, and the Averaging method for regression problems. Thus, we can explore the other combination strategies in the future, which may further improve the effect of the model.

# REFERENCES

[Zhou, 2012] Z.-H. Zhou. Ensemble Methods: Foundations and Algorithms. CRC, Boca Raton, FL, 2012.

[Breiman, 2001] L. Breiman. Random forests. Machine learning, 45(1):5–32, 2001.

[Wolpert, 1992] Wolpert, D. H. (1992). Stacked generalization. Neural networks, 5(2), 241-259.

[Breiman, 1996] L. Breiman. Bagging predictors. Machine Learning, 24(2):123–140, 1996.

[Chen, 2016] T.-Q. Chen and C. Guestrin. XGBoost: A scalable tree boosting system. In KDD, pages 785–794, 2016.

[Schapire, 1990] Schapire, R. E. (1990). The strength of weak learnability. Machine learning, 5(2), 197-227.

[Zhou, 2017] Z.-H. Zhou and J. Feng. Deep forest: Towards an alternative to deep neural networks. In IJCAI, pages 3553–3559, 2017.

[Tannor, 2019] Philip Tannor and Lior Rokach. AugBoost: Gradient Boosting Enhanced with Step-Wise Feature Augmentation. In IJCAI, pages 3555-3561, 2019.

[Zhang and Zhang, 2008] Chun-Xia Zhang and Jiang-She Zhang. Rotboost: A technique for combining rotation forest and adaboost. Pattern recognition letters, 29(10):1524–1536, 2008.

[Alcala-Fdez, 2011] Jesus Alcala-Fdez, Alberto Fernandez, Julian Luengo, Joaquin Derrac, and Salvador Garcia. Keel data-mining software tool: Data set repository, integration of algorithms and experimental analysis framework. Multiple-Valued Logic and Soft Computing, 17(2-3):255–287, 2011.

[Dheeru and Karra Taniskidou, 2017] Dua Dheeru and EfiKarra Taniskidou. UCI machine learning repository, 2017.

# APPENDIX

Pseudo Code:

---
**Algorithm 1** Train AGM-v3
---
**Input:** training set{X, y}, Base Model Sets $\{H_i(x)\}_1^N$, Model $M$, val-model $H$
**Output:** model AGM-v3
 1: Initialize $Acc_w \leftarrow 0, Acc_d \leftarrow [\ ], n \leftarrow 0, maxnum \leftarrow 3, Models \leftarrow \{\}$
 2: **function** GET WIDTH({X, y}, Model $M$, val-model $H$)
 3:     $i \leftarrow 0$
 4:     **while** True **do**
 5:         Train $M_i(x)$ on {X, y}
 6:         New-feature $\leftarrow$ Predict X by $M_i(x)$
 7:         X $\leftarrow$ Concatenate X and New-feature
 8:         Train $H$ on {X, y} and obtain $acc$
 9:         **if** $acc > Acc_w$ **then** $Acc_w \leftarrow acc, i++$
10:         **else break**
11:         **end if**
12:     **end while**
13:     **return** $i$
14: **end function**
15:
16: X-train, X-val, y-train, y-val $\leftarrow \{X, y\}$
17: **while** True **do**
18:     $num \leftarrow$ **function** Get Width
19:     **if** $n$ *is not* 0 **then**
20:         X-train $\leftarrow$ concatenate(X-train, new-train-feature)
21:         X-val $\leftarrow$ concatenate(X-val, new-val-feature)
22:         $PCA$ on X-train, X-val
23:     **end if**
24:     **for** $i = 0 \rightarrow num$ **do**
25:         Get one model from Base Model Stes randomly
26:         Train the model on {X-train, y-train}
27:         new-train-feature, new-val-feature $\leftarrow$ Predict X-train, X-val by the model
28:         Add the model to Models
29:     **end for**
30:     Evaluate Models on training set to obtain $acc$
31:     $Acc_d[n] \leftarrow acc$
32:     **if** $n = 0$ **then**
33:         $n++$, **continue**
34:     **end if**
35:     **if** $Acc_d[n] < Acc_d[n-1]$ **then break**
36:     **else**
37:         **if** $Acc_d[n] = Acc_d[n-1]$ **then**
38:             $maxnum--$
39:             **if** $maxnum = 0$ **then break**
40:             **end if**
41:             $n++$, **continue**
42:         **else** $n++$, **continue**
43:         **end if**
44:     **end if**
45: **end while**
46: **return** $Models$